Г. А. Чураков
# ЧАСТЕРЕЧНАЯ РАЗМЕТКА ДЛЯ ВЫДЕЛЕНИЯ СКЕЛЕТНОЙ СТРУКТУРЫ ПРЕДЛОЖЕНИЙ

В статье описан процесс разработки модели для нанесения частеречной аннотации на текст с использованием переноса обучения BERT. Процесс подготовки данных и оценка полученных результатов. Выявлено, что предложенный способ позволяет достичь хороших результатов в нанесении разметки на текст.

**Ключевые слова:** частеречная разметка, морфологический анализ, обработка естественного языка.

G. A. Churakov
# POS-TAGGING TO HIGHLIGHTTHE SKELETAL STRUCTURE OF SENTENCES

The article describes the process of developing a model for applying partial annotation to text using BERT learning transfer. Process of data preparation and evaluation of obtained results. It has been found that the proposed method makes it possible to achieve good results in marking text.

## Введение

Одним из подвидов задачи по извлечению структурированной информации из текста является задача по частеречной разметке (Part-of-Speech Tagging), процесс аннотирования слова в тексте (корпусе) определенной части речи в зависимости от ее определения и контекста.

Разметка частей речи представляет собой более сложную задачу, чем просто составление списка слов с соответствующими частями речи. Это связано с тем, что некоторые слова могут включать в себя несколько частей речи в зависимости от контекста, а некоторые части речи сложны по своей природе. Это нередкое явление в естественных языках, где значительная часть словоформ, в отличие от многих искусственных языков, неоднозначна.

Данная задача решалась различными методами, первые подходы к решению данного вопроса основывались на составлении правил вручную. Это требовало обширных знаний о грамматике конкретного языка, что делало подобную систему аннотирования трудно переносимой на другие языки. В данной статье будет рассмотрен подход с использованием нейросетевой модели использующей архитектуру BERT в качестве основы.

**Объектом исследования** служат синтаксические и грамматические признаки, по которым данные лексемы можно распознать в тексте.

**Предметом исследования** является возможность применения алгоритма автоматического аннотирования текстовых данных на русском языке частеречной разметкой.

**Цель** – разработать модель, способную с хорошей точностью выделять из текста на русском языке скелетную частеречевую структуру.

**Материалом исследования** является корпус, из 100 предложений на русском языке, размеченных экспертом.

**Методы исследования**: интуитивное выделение необходимых признаков для определения отнесенности слова части речи – представление слов в закодированном виде, моделирование синтаксического уровня при представлении предложения в виде закодированной последовательности, аугментация текстовых данных.

**Результатом работы** является модель, позволяющая распознавать в тексте части речи. Практическая значимость заключается в том, что тегирование текста частями речи является важным компонентом на этапе подготовки данных. Отсутствие этой лингвистической аннотации затрудняет дальнейший анализ текста, в первую очередь из-за неоднозначности значений слов. Конкретно эта модель предназначения для аннотирования текста базовыми тегами для улучшения качества машинного перевода с применением скелетных структур [10,13,14,16,18].

## Метрики оценки качества разметки

Сформулировав задачу POS-тэггинга, как многоклассовую классификацию, использовались соответствующие метрики. Учитывая также природу языка следует заметить дисбаланс классов в выборке, в этих условиях в качестве основной метрики была выбрана взвешенная F-1 мера. Данная метрика в задаче многоклассовой классификации вычисляется с точки зрения «один-против всех», когда к True Positive - относятся все классифицированные токены, а объекты других классов относятся к False Positive. Формула ниже описывает вычисление F1 меры для отдельно взятого класса.

$$\text{F-1 Score(class=a)} = \frac{2 * \text{TP (class=a)}}{2 * \text{TP(class=a)} + \text{FP (class=a)} + \text{FN (class=a)}}$$

*Формула 1 – Вычисление F-1 меры.*

В задаче классификации с несколькими классами используется взвешеная сумма оценок для каждого из классов.

## Обработка данных

Корпус размеченного текста представлял собой 100 предложений, он был предварительно предобработан, данные были трансформированы в последовательности типа: предложение – набор тэгов для каждого слова в предложении. Такого объема данных недостаточно для обучения модели, поэтому были предприняты шаги по увеличению выборки[7]. Предложения тестовой выборки были нарезаны на фрагменты при помощи скользящего окна размером [1; количество слов в предложении]. Таким образом, количество наблюдений увеличилось тренировочную выборку в до 20000 наблюдений.

## Архитектура модели

Для построения модели данные недостоточно разнообразны, поэтотму было решение произвести перенос обучения.

В качестве базовой модели решения выбрана RuBERT-base[9] содержащая в себе енкодер блоки сети трансформер[3,4,5]. Среди особенностей модели хочется выделить ее устройство:

- предобученый Токенизатор – BPE[1].

- Модель обучена на задаче маскированного языкового моделирования, в процессе обучения модель извлекает из последовательности фиксированной длинны признаки необходимые для предсказания скрытого токена, расположенного на случайной позиции в последовательности.

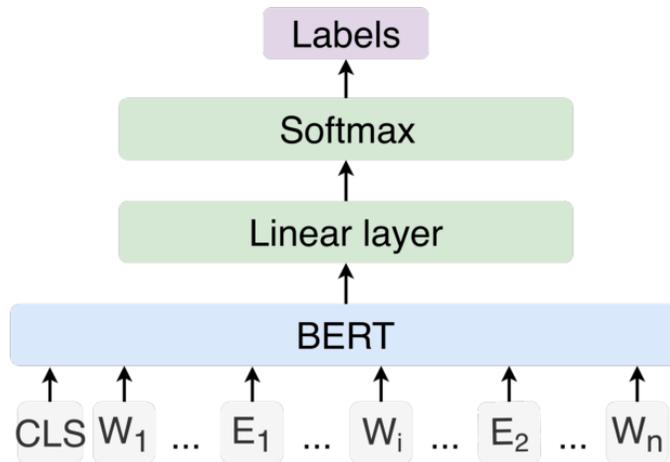

Рисунок 1– Картинка из оригинальной статьи[17]: дообучение BERT для задачи Token Classification.

Обучение модели

Как описано в оригинальной статье, BERT можно легко дообучить под решение узкой задачи, для этого добавим к предобученой русскоязычной модели полносвязный слой и определим функцию активации Softmax, на выходе модели получим закодированные значения классов токенов, которые затем раскодируем во время постобработки.

Далее обучим модель на задачу Token Classification с учителем, будем оценивать качество после каждой эпохи и при помощи оптимизатора Adam корректировать веса модели методом обратного распространения ошибки.

Процесс обучения модели можно наблюдать на рисунке ниже.

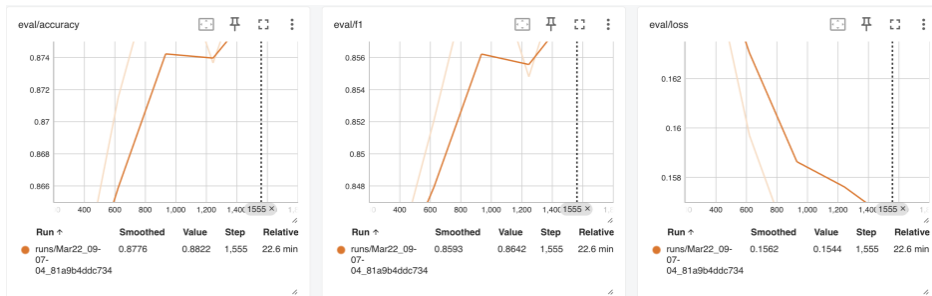

На валидации были получены следующие метрики для модели:

- **F1: 0.8642**

- **Accuracy: 0.8822**

## Итоги

Целью данной работы разработка модели, способной с хорошей точностью выделять из текста на русском языке скелетную частеречевую структуру. Важно признать, что обучение и оценка модели проводились на относительно ограниченном наборе данных, поэтому результаты могут отличаться в зависимости от набора данных.

Данную работу можно продолжать за счёт увеличения объёма тренировочных данных (мультиязычность, более сбалансированная выборка) и подбора гиперпараметров нейронной сети. В ходе анализа результатов работы модели, также было выявлено, что модель показывает способна к разметке на другом языке, корректно выделяются части речи, которых встретилось сравнительно много в наборе данных для обучения.

С результатами работы можно ознакомиться на следующих ресурсах:

- Github: https://github.com/disk0Dancer/rubert-finetuned-pos.git;

- Hugging Face Demo: https://huggingface.co/disk0dancer/ruBert-base-finetuned-pos.

## Библиография

1. Sennrich R., Haddow B., Birch A. Neural Machine Translation of Rare Words with Subword Units // 2016.

2. Bojanowski P. [и др.]. Enriching Word Vectors with Subword Information // 2017.

3. Vaswani A. [и др.]. Attention Is All You Need // 2023.

4. Devlin J. [и др.]. BERT: Pre-training of Deep Bidirectional Transformers for Language Understanding // 2019.

5. Tenney I., Das D., Pavlick E. BERT Rediscovers the Classical NLP Pipeline // 2019.

6. Wang C., Cho K., Gu J. Neural Machine Translation with Byte-Level Subwords // 2019.

7. Thakur N. [и др.]. Augmented SBERT: Data Augmentation Method for Improving Bi-Encoders for Pairwise Sentence Scoring Tasks // 2021.

8. Hui W. Performance of Transfer Learning Model vs. Traditional Neural Network in Low System Resource Environment // 2020.

9. Zmitrovich D. [и др.]. A Family of Pretrained Transformer Language Models for Russian // 2023.

10. Ailamazyan Program Systems Institute of RAS, Pereslavl-Zalessky, 152020, Russian Federation, Trofimov I. V. Automatic Morphological Analysis for Russian: Application-Oriented Survey // PROGRAMMNAYA INGENERIA. 2019. № 9–10 (10). C. 391–399.

11. Hiroki Nakayama Seqeval: A Python framework for sequence labeling evaluation [Электронный ресурс]. URL: https://github.com/chakki-works/seqeval.

12. Liao W., Veeramachaneni S. A simple semi-supervised algorithm for named entity recognition Boulder, Colorado: Association for Computational Linguistics, 2009. C. 58–65.

13. Mylnikova A. V., Trusov V. A., Mylnikov L. A. Use of Text Skeleton Structures for the Development of Semantic Search Methods // Automatic Documentation and Mathematical Linguistics. 2023. № 5 (57). C. 301–307.

14. Novikova A. Direct Machine Translation and Formalization Issues of Language Structures and Their Matches by Automated Machine Translation for the Russian-English Language Pair 2018.

15. Pennington J., Socher R., Manning C. Glove: Global Vectors for Word Representation Doha, Qatar: Association for Computational Linguistics, 2014. C. 1532–1543.

16. Sapin A. S. Building neural network models for morphological and morpheme analysis of texts // Proceedings of the Institute for System Programming of the RAS. 2021. № 4 (33). C. 117–130.

17. Wei Q. [и др.]. Relation Extraction from Clinical Narratives Using Pre-trained Language Models // AMIA ... Annual Symposium proceedings. AMIA Symposium. 2019. (2019). C. 1236–1245.

18. Zhang Y., Jin R., Zhou Z.-H. Understanding bag-of-words model: a statistical framework // International Journal of Machine Learning and Cybernetics. 2010. № 1–4 (1). C. 43–52.

19. INTELLECTUAL ANALYSIS OF DATA ON THE BASIS OF STANFORD CoreNLP FOR POS TAGGING OF TEXTS IN THE RUSSIAN LANGUAGE // Systems and Means of Informatics. 2018.

20. Classical ML Equations in LaTeX [Электронный ресурс]. URL: https://blmoistawinde.github.io/ml_equations_latex/.

21. MLM [Электронный ресурс]. URL: https://www.sbert.net/examples/unsupervised_learning/MLM/README.html.

22. Summary of the tokenizers [Электронный ресурс]. URL: https://huggingface.co/docs/transformers/tokenizer_summary.

23. what-is-word2vec-and-how-does-it-work [Электронный ресурс]. URL: https://swimm.io/learn/large-language-models/what-is-word2vec-and-how-does-it-work.

## Об авторе

**Григорий Александрович Чураков** – студент бакалавриата образовательной программы «Программная инженерия», факультета социально-экономических и компьютерных наук, Пермский филиал Национального исследовательского университета «Высшая школа экономики», Россия, Пермь, email: gachurakov@edu.hse.ru